%% file: 2026_IEEE_IV_learned3DNMS.tex
\documentclass[conference]{IEEEtran}
\IEEEoverridecommandlockouts{}
\input{packages.tex}

\def\BibTeX{{\rm B\kern-.05em{\sc i\kern-.025em b}\kern-.08em
    T\kern-.1667em\lower.7ex\hbox{E}\kern-.125emX}}

\newtoggle{todos}
\settoggle{todos}{false}
\newif\ifdraft\draftfalse{}

\begin{document}
\graphicspath{{figures/}}

\newcommand\copyrighttext{%
	\footnotesize \textcopyright 2026 IEEE. Personal use of this material is permitted.  Permission from IEEE must be obtained for all other uses, in any current or future media, including reprinting/republishing this material for advertising or promotional purposes, creating new collective works, for resale or redistribution to servers or lists, or reuse of any copyrighted component of this work in other works. This work has been accepted for publication in the proceedings of the IEEE Intelligent Vehicles Symposium 2026.}
\newcommand\copyrightnotice{%
	\begin{tikzpicture}[remember picture,overlay]
	\node[anchor=south,xshift=5pt,yshift=10pt] at (current page.south) {\fbox{\parbox{\dimexpr\textwidth-\fboxsep-\fboxrule\relax}{\copyrighttext}}};
	\end{tikzpicture}%
}

\title{Learned Non-Maximum Suppression for 3D Object Detection}

\author{Timo Osterburg, Stefan Schütte, and Torsten Bertram\\
\small Institute of Control Theory and Systems Engineering, TU Dortmund University, Germany.

}


    

\maketitle
\copyrightnotice
\begin{abstract}
Post-processing is a critical stage in LiDAR-based 3D object detection, where dense and overlapping proposals must be filtered for compact and reliable perception. 
This work introduces two learned filtering modules that replace heuristic non-maximum suppression (NMS) by leveraging relations among detections. 
D2D-Rescore employs transformer-based detection-to-detection (D2D) attention, while GossipNet3D adapts the 2D GossipNet concept to 3D through localized message passing in bird's-eye view. 
A metric-aware matching strategy aligned with the nuScenes evaluation protocol ensures consistent training and validation behavior, improving overall detection performance.
Both approaches improve \gls{map}, \gls{nds}, and true positive quality compared to CircleNMS, particularly for small and infrequent classes, while adding minimal computational overhead.
These results demonstrate that learned, detection-level filtering can enhance 3D detector reliability without modifying the base network, offering a principled alternative to heuristic suppression. Code is available at \url{https://github.com/rst-tu-dortmund/learned-3d-nms}.
\end{abstract}


\input{symbols/symbols.tex}
\input{glos.tex}

\input{sections/introduction.tex}
\input{sections/related_work.tex}
\input{sections/problem_formulation.tex}
\input{sections/methodology.tex}
\input{sections/evaluation.tex}

\input{sections/conclusion_and_outlook.tex}

\printbibliography



\end{document}

%% file: packages.tex
\usepackage{etoolbox} 
\usepackage{graphicx}
\usepackage[table]{xcolor}
\usepackage[utf8]{inputenc}
\usepackage{mathptmx}
\usepackage{array,contour}
\usepackage{tabularx}
\usepackage{booktabs}
\usepackage{multirow}
\usepackage{ragged2e}
\usepackage[singlelinecheck=false]{caption}
\usepackage[T1]{fontenc}

\usepackage{parskip}
\usepackage{amsmath, amssymb, amsbsy, bm}
\usepackage{amsfonts}
\usepackage{amsthm}
\usepackage{mathtools}
\usepackage{glossaries}
\DeclareMathOperator*{\argmax}{argmax}
\usepackage{float}
\usepackage{siunitx}

\usepackage{tikz, pgfplots}
\usetikzlibrary{spy}
\usetikzlibrary{backgrounds}
\usetikzlibrary{decorations}
\usetikzlibrary{shapes,snakes}
\usetikzlibrary{decorations.markings}
\usepgfplotslibrary{groupplots}
\usetikzlibrary{shapes.geometric}
\usetikzlibrary{calc}
\usetikzlibrary{positioning}
\usetikzlibrary{fit}
\usetikzlibrary{matrix}
\usepackage{tikzscale}

\usepackage{grffile}
\usepackage{currfile}
\usepackage[justification=centering]{caption}
\usepackage{subcaption}

\usepackage[english]{babel}
\usepackage{pgfplots}
\usepackage{enumitem}   

\usepgfplotslibrary{groupplots,dateplot}
\usetikzlibrary{positioning,arrows.meta}   
\usepgflibrary{patterns}
\pgfplotsset{compat=newest}
\usepackage[backend=biber,style=ieee,maxbibnames=3]{biblatex}

\makeglossaries
\usepackage[mathcal]{euscript}

%% file: symbols/symbols.tex
\input{symbols/detections/sets.tex}
\input{symbols/notation.tex}
\input{symbols/detections/variables.tex}
\input{symbols/method/naming.tex}

%% file: symbols/detections/sets.tex
\newcommand{\SetDenseDets}{
    \mathcal{D}_{\textrm{d}}
}

\newcommand{\SetFilteredDets}{
    \mathcal{D}_{\textrm{f}}
}

\newcommand{\SetAnnDets}{
    \mathcal{G}
}

%% file: symbols/notation.tex
\newcommand{\Vect}[1]{  
    \mathbf{#1}
}

%% file: symbols/detections/variables.tex
\newcommand{\locX}{
    x
}
\newcommand{\locY}{
    y
}
\newcommand{\locZ}{
    z
}

\newcommand{\extendX}{
    l
}

\newcommand{\extendY}{
    w
}

\newcommand{\extendZ}{
    h
}

\newcommand{\heading}{
    \theta
}

\newcommand{\classID}{
    c_{\textrm{id}}
}

\newcommand{\score}{
    s
}

\newcommand{\velX}{
    v_x
}

\newcommand{\velY}{
    v_y
}

\newcommand{\velZ}{
    v_z
}

%% file: symbols/method/naming.tex
\newcommand{\ApproachName}{D2D-Rescore}

%% file: glos.tex
\newacronym{bev}{BEV}{bird-eye-view}
\newacronym{iou}{IoU}{intersection-over-union}
\newacronym{nms}{NMS}{non-maximum suppression}
\newacronym{rois}{RoIs}{regions of interest}
\newacronym{map}{mAP}{mean Average Precision}
\newacronym{nds}{NDS}{nuScenes Detection Score}
\newacronym{tps}{TP}{True Positives}

%% file: sections/introduction.tex
\section{Introduction}\label{sec:introduction}

Reliable 3D object detection is essential for autonomous driving, forming the foundation for downstream tasks such as tracking, motion prediction, and planning.  
LiDAR-based detectors, including CenterPoint~\cite{yin2021center}, PointPillars~\cite{lang2019pointpillars}, and DSVT~\cite{wang2023dsvt}, produce dense sets of detection candidates from spatial priors such as anchors or center points, resulting in numerous overlapping proposals per frame.
To obtain a compact and reliable perception of the environment, these proposals must be filtered to remove duplicates and false positives.

Classical post-processing performs this filtering through \gls{nms} and confidence thresholding.  
Although efficient, these heuristics are non-differentiable, rely on hand-tuned thresholds, and ignore contextual relations among detections.  
They therefore lack adaptation to scene-specific conditions such as occlusion or object density, which can suppress valid detections or retain false positives in crowded scenes \cite{bodla2017soft}.

The Detection Transformer (DETR)~\cite{carion2020end} introduced a training objective that integrates suppression into the detector.
While its extensions to 3D detection~\cite{wang2022detr3d, erabati2023li3detr} adopt this paradigm, they focus on introducing new transformer-based architectures with increased memory requirements and inference latency, limiting their real-time applicability in automotive perception.

Several learned filtering methods in 2D detection have shown that context-aware score refinement can surpass heuristic suppression~\cite{hosang2017learning, hu2018relation, ding2022end}.
In contrast, 3D detection has seen limited exploration of such approaches, motivating an investigation into whether existing detectors can profit from learned filtering without architectural or training modifications.

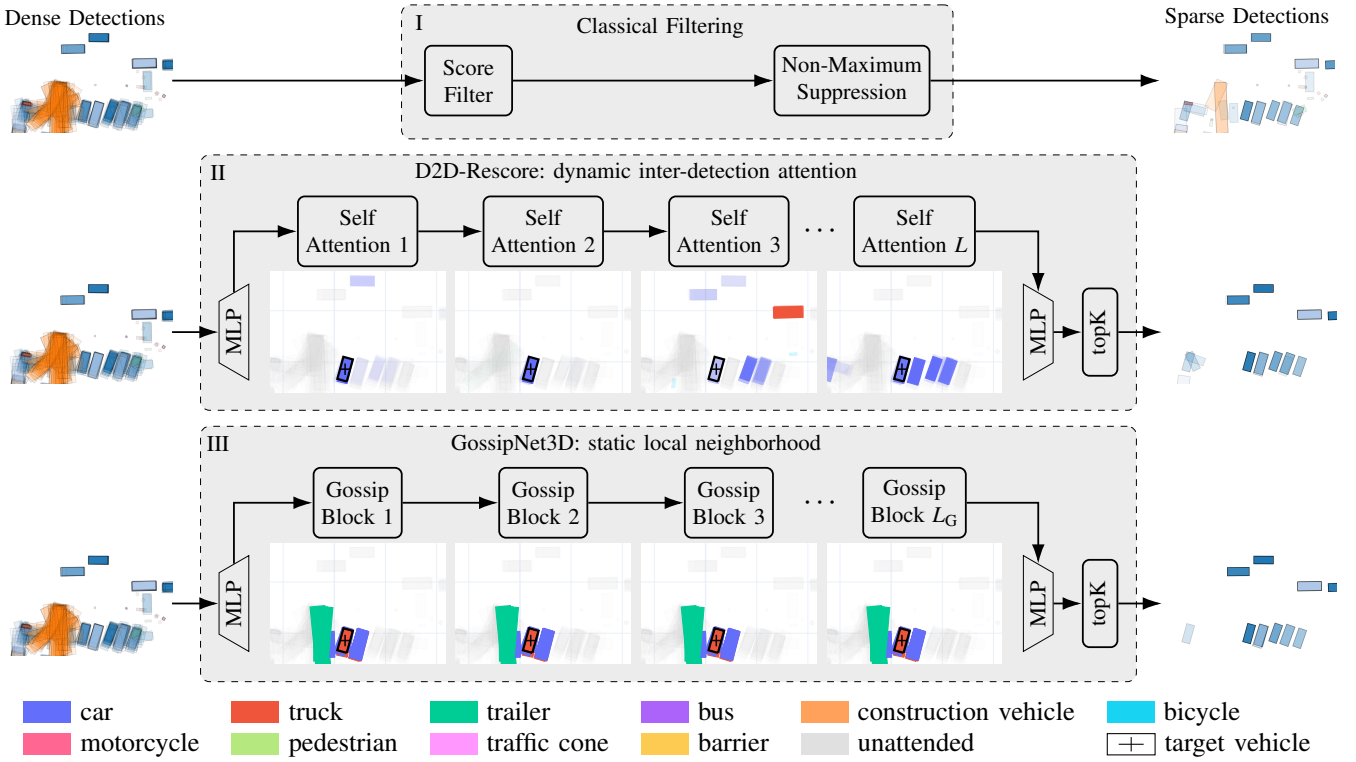
\begin{figure*}
    \vspace{2pt}
    \centering

    \resizebox{\textwidth}{!}{
    \input{figures/pipelines/classical_filtering_pipeline.tex}
    }
    \par\vspace*{0.2cm}
    \resizebox{\textwidth}{!}{
    \input{figures/pipelines/d2d_rescore.tex}
    }
    \par\vspace*{0.2cm}
    \resizebox{\textwidth}{!}{
    \input{figures/pipelines/gossipnet.tex}
    }
    \par\vspace*{0.2cm}
    
    \input{figures/attention_viz_legend.tex}

    \caption{Overview of the post-processing pipelines: classical filtering (I) uses score thresholding and non-maximum suppression, while the proposed D2D-Rescore (II) and GossipNet3D (III) refine dense detector outputs via learnable layers and top-$K$ selection to yield improved sparse detections. D2D-Rescores attention adapts in each layer and scene based on detection features, whereas GossipNet3D relies on static local neighborhoods defined by spatial proximity.}
    \label{fig:overview}
\end{figure*}

This work addresses this by introducing lightweight, detector-agnostic learned filtering modules operating directly on 3D detection sets.  
It contributes by:
\begin{itemize}[noitemsep,topsep=0pt]
  \item introducing D2D-Rescore, a transformer-based module that models inter-detection relations via self-attention to refine confidence scores,
  \item proposing GossipNet3D, the first adaptation of GossipNet~\cite{hosang2017learning} to 3D detection, aggregating features within local \gls{bev} neighborhoods, and
  \item demonstrating consistent improvements in \gls{map}, \gls{nds}, and true-positive quality on nuScenes, with notable gains for rare and small classes.
\end{itemize}
Both modules act as detector-agnostic post-processing components that replace heuristic \gls{nms} in the 3D detection pipeline, as illustrated in Figure~\ref{fig:overview}.

%% file: figures/pipelines/classical_filtering_pipeline.tex
\tikzset{
  block/.style   = {draw, thick, rounded corners=3pt, minimum width=0.075\linewidth,
                    minimum height=0.06\linewidth, align=center},
  img/.style     = {inner sep=0pt, outer sep=0pt},
  >={Latex[length=3mm,width=2mm]}
}

\begin{tikzpicture}[node distance=0.06\linewidth, transform shape]

\def\nodesep{0.06\linewidth}
\def\rowgap{0.03\linewidth}   
\def\imw{0.2\linewidth}      
\def\imh{0.15\linewidth}      
\def\trimvalv{3.5cm}      
\def\trimvall{1.8cm}      
\def\trimvalr{2.8cm}      

\coordinate (origin) at (0.0, 0);
\node[img, anchor=center, rotate=90, transform shape] (input) at (0.0, 0) {\includegraphics[height=\imh, trim=4.2cm 5cm 10cm 5cm, clip]{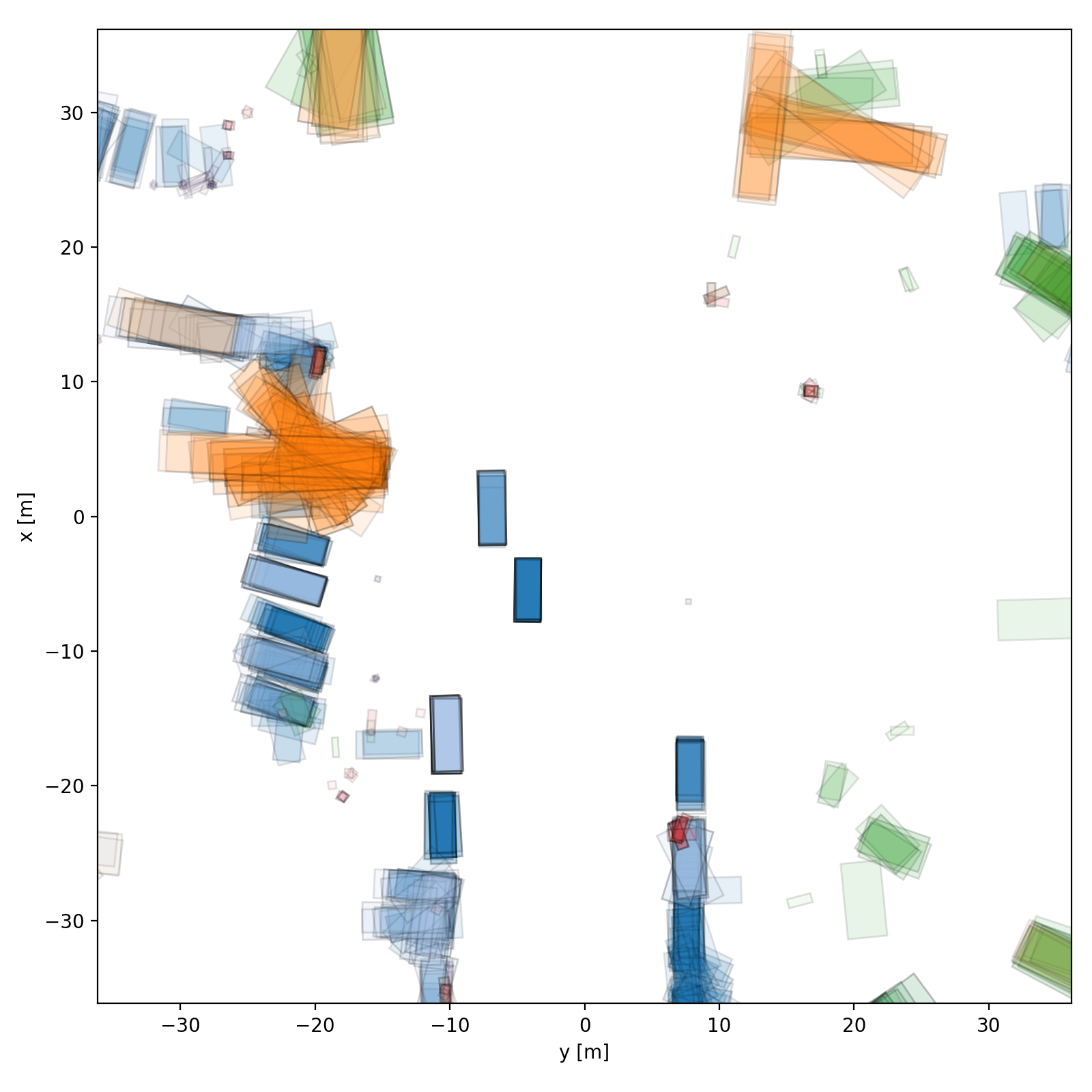}};
\node[] (inlbl) at ($(input.east)+(0,5pt)$) {Dense Detections};

\node[img, anchor=center, rotate=90, transform shape] (out) at (\linewidth, 0) {\includegraphics[height=\imh, trim=4.2cm 5cm 10cm 5cm, clip]{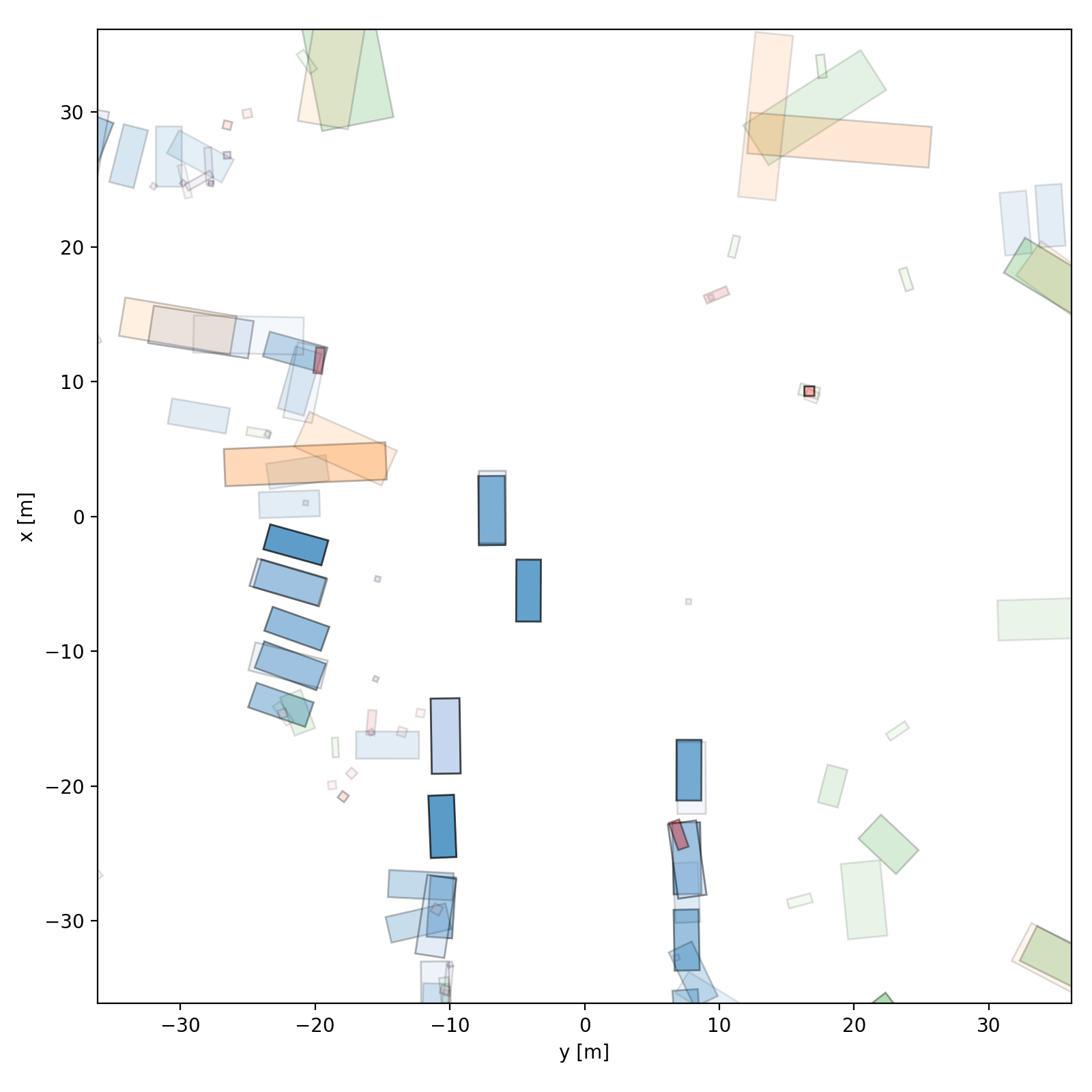}};

\node[block, anchor=center] (score) at ($(input)!0.33!(out)$) {Score\\Filter};
\node[] (inlbl) at ($(out.east)+(0,5pt)$) {Sparse Detections};

\node[block, anchor=center] (nms) at ($(input)!0.66!(out)$) {Non-Maximum\\Suppression};

\node[] (cflbl) at ($(score.north)!0.5!(nms.north)+(0,0.015\linewidth)$) {Classical Filtering};

\begin{pgfonlayer}{background}
  \node[fill=gray!15, draw, dashed, rounded corners,
        fit=(cflbl.west) (score) (nms),
        inner sep=10pt] (background) {};
\end{pgfonlayer}

\node[] (apprlbl) at ($(background.west |- background.north)+(8pt, -8pt)$) {I};

\draw[->, thick] (input) -- (score);
\draw[->, thick] (score) -- (nms);
\draw[->, thick] (nms) -- (out);

\end{tikzpicture}

%% file: figures/pipelines/d2d_rescore.tex
\tikzset{
  block/.style   = {draw, thick, rounded corners=3pt, minimum width=0.075\linewidth,
                    minimum height=0.06\linewidth, align=center},
  img/.style     = {inner sep=0pt, outer sep=0pt},
  >={Latex[length=3mm,width=2mm]}
}

\begin{tikzpicture}[node distance=0.06\linewidth, transform shape]

\def\nodesep{0.06\linewidth}
\def\rowgap{0.03\linewidth}   
\def\imw{0.2\linewidth}      
\def\imh{0.15\linewidth}      
\def\trimvalv{3.5cm}      
\def\trimvall{1.8cm}      
\def\trimvalr{2.8cm}      

\node[img, anchor=center, rotate=90, transform shape] (input) at (0, 0) {\includegraphics[height=\imh, trim=4.2cm 5cm 10cm 5cm, clip]{dense_bev_no_title_small.png}};

\node[img, anchor=center, rotate=90, transform shape] (out) at (\linewidth, 0) {\includegraphics[height=\imh, trim=4.2cm 5cm 10cm 5cm, clip]{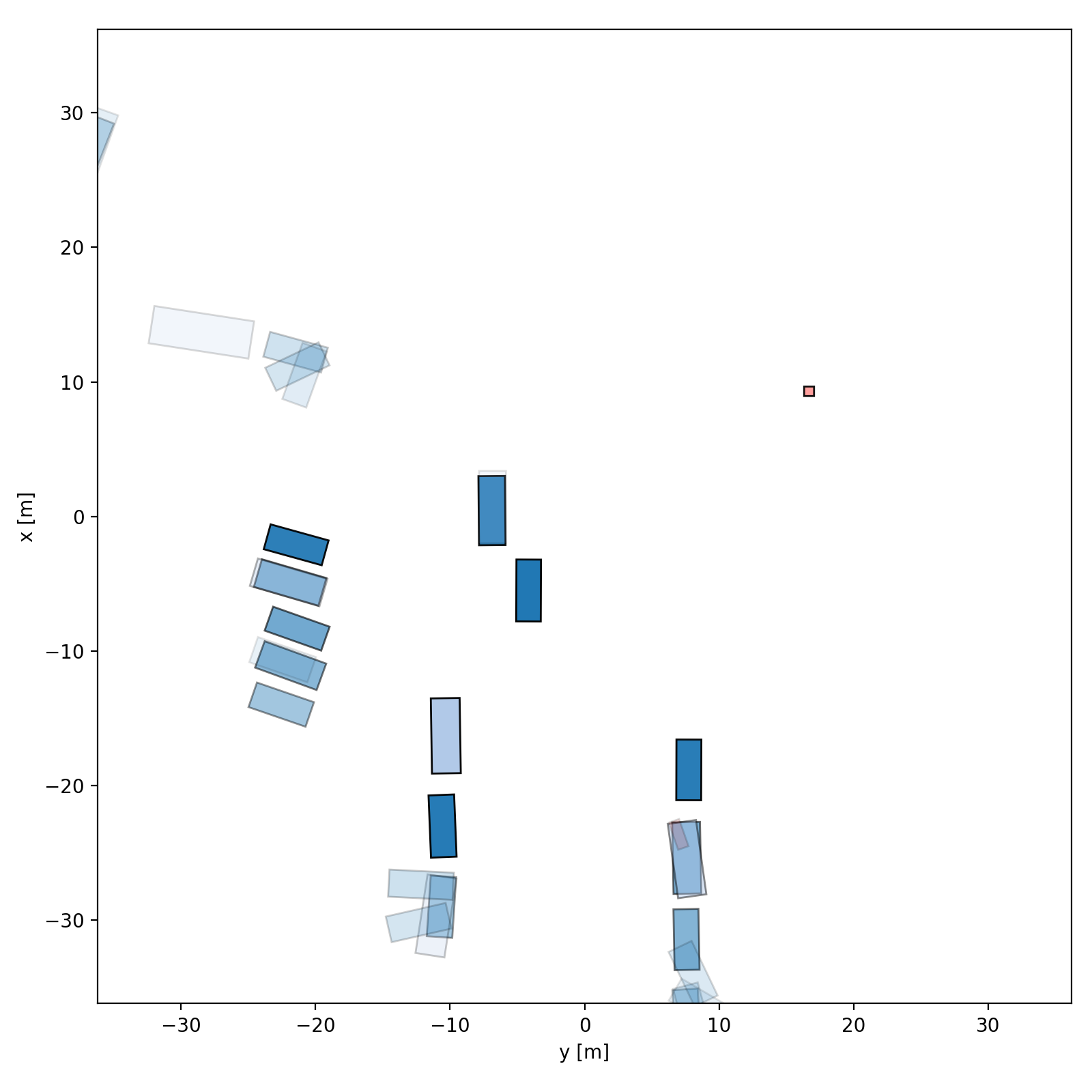}};

\coordinate (left_anchor) at (input.south);
\coordinate (right_anchor) at (out.north);

\node[trapezium, trapezium stretches=true, draw, rotate=90, transform shape] (in_mlp) at ($(left_anchor)!1/16!(right_anchor)$) {MLP};

\node[trapezium, trapezium stretches=true, draw, rotate=90, shape border rotate=180, transform shape] (out_mlp) at ($(left_anchor)!14/16!(right_anchor)$) {MLP};

\node[block, rotate=90, transform shape, minimum height=0.02\linewidth] (topk) at ($(left_anchor)!15/16!(right_anchor)$) {topK};

\node[img, rotate=90, transform shape] (d2d_attn_1) at ($(left_anchor)!3/16!(right_anchor)$)
  {\includegraphics[height=\imh, trim={\trimvall} {\trimvalv} {\trimvalr} {\trimvalv}, clip]{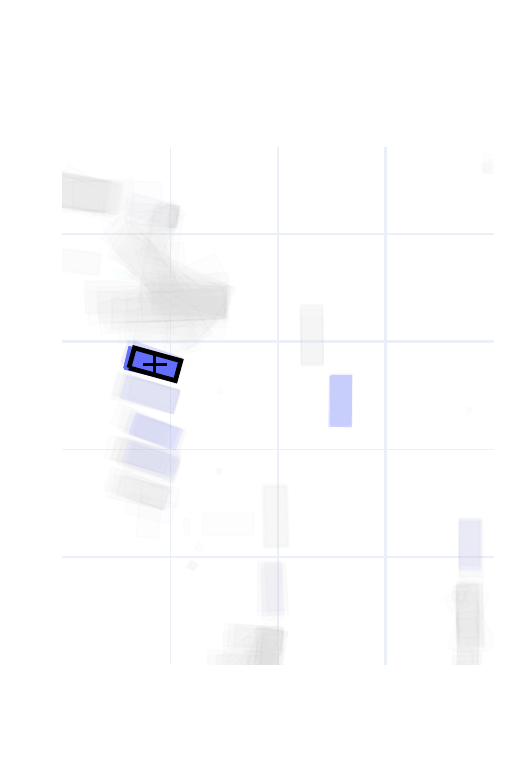}};

\node[img, rotate=90] (d2d_attn_2) at ($(left_anchor)!6/16!(right_anchor)$)
  {\includegraphics[height=\imh, trim={\trimvall} {\trimvalv} {\trimvalr} {\trimvalv}, clip]{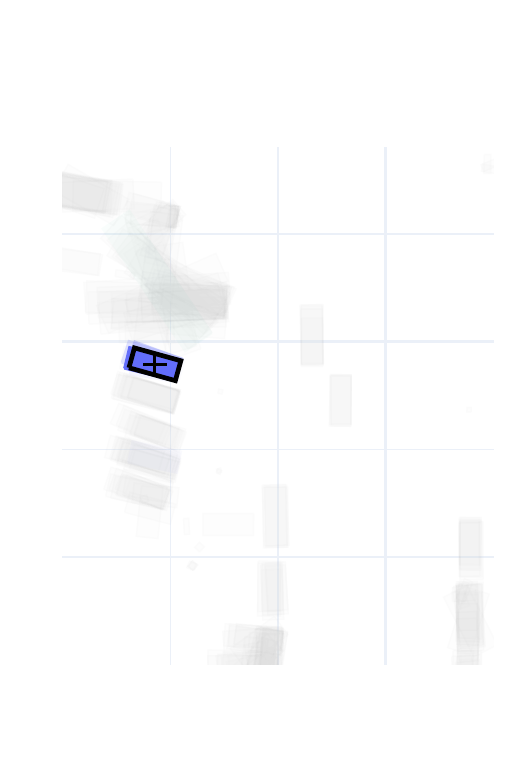}};

\node[img, rotate=90] (d2d_attn_3) at ($(left_anchor)!9/16!(right_anchor)$)
  {\includegraphics[height=\imh, trim={\trimvall} {\trimvalv} {\trimvalr} {\trimvalv}, clip]{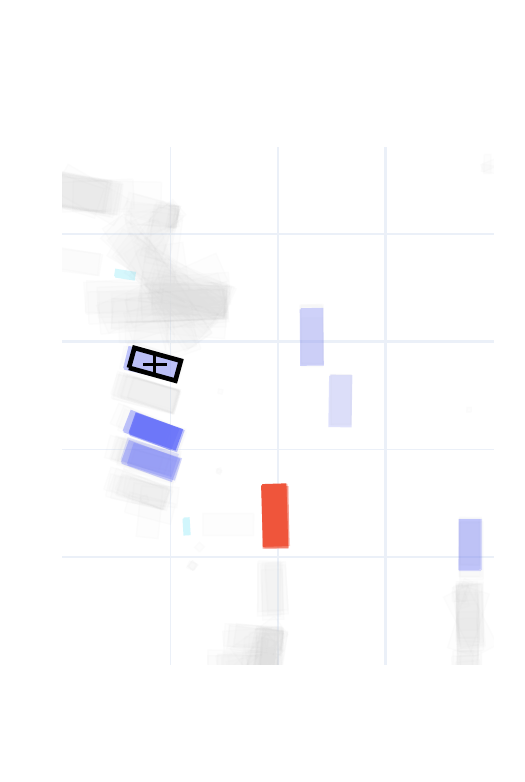}};

\node[img, rotate=90] (d2d_attn_4) at ($(left_anchor)!12/16!(right_anchor)$)
  {\includegraphics[height=\imh, trim={\trimvall} {\trimvalv} {\trimvalr} {\trimvalv}, clip]{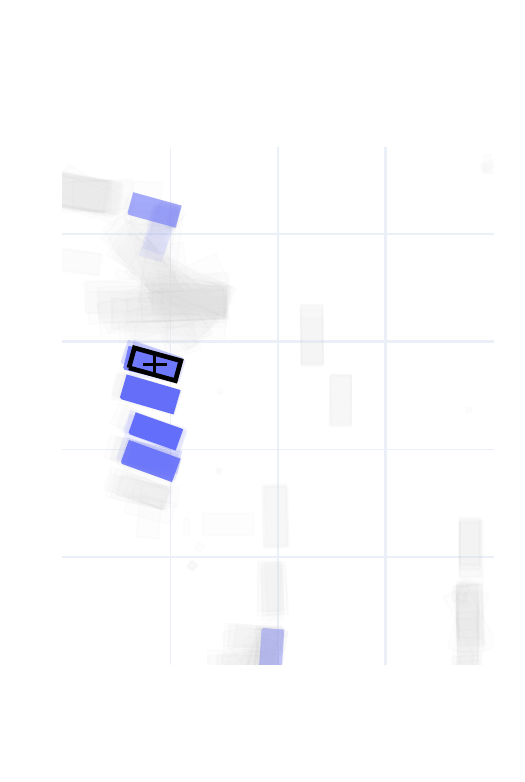}};

\node[block, anchor=center] (d2dlayer1) at ($(d2d_attn_1.east)+(0,18pt)$) {Self\\Attention 1};
\node[block, anchor=center] (d2dlayer2) at ($(d2d_attn_2.east)+(0,18pt)$) {Self\\Attention 2};
\node[block, anchor=center] (d2dlayer3) at ($(d2d_attn_3.east)+(0,18pt)$) {Self\\Attention 3};
\node[block, anchor=center] (d2dlayer4) at ($(d2d_attn_4.east)+(0,18pt)$) {Self\\Attention $L$};

\node[anchor=center] (dots) at ($(d2dlayer3)!0.5!(d2dlayer4)$) {\Large $\cdots$};

\node[] (d2dlbl) at ($(d2dlayer2.north)!0.5!(d2dlayer3.north)+ (0,10pt)$) {D2D-Rescore: dynamic inter-detection attention};

\begin{pgfonlayer}{background}
  \node[fill=gray!15, draw, dashed, rounded corners,
        fit=(in_mlp) (topk) (d2dlayer1) (d2d_attn_1) (d2dlbl.west),
        inner sep=8pt] (background) {};
\end{pgfonlayer}

\node[] (apprlbl) at ($(background.west |- background.north)+(8pt, -8pt)$) {II};

\draw[->, thick] (input) -- (in_mlp);
\draw[->, thick] (in_mlp) |- (d2dlayer1);
\draw[->, thick] (d2dlayer1) -- (d2dlayer2);
\draw[->, thick] (d2dlayer2) -- (d2dlayer3);
\draw[->, thick] (d2dlayer4) -| (out_mlp);
\draw[->, thick] (out_mlp) -- (topk);
\draw[->, thick] (topk) -- (out);

\end{tikzpicture}

%% file: figures/pipelines/gossipnet.tex
\tikzset{
  block/.style   = {draw, thick, rounded corners=3pt, minimum width=0.075\linewidth,
                    minimum height=0.06\linewidth, align=center},
  img/.style     = {inner sep=0pt, outer sep=0pt},
  >={Latex[length=3mm,width=2mm]}
}

\begin{tikzpicture}[node distance=0.06\linewidth, transform shape]

\def\nodesep{0.06\linewidth}
\def\rowgap{0.03\linewidth}   
\def\imw{0.2\linewidth}      
\def\imh{0.15\linewidth}      
\def\trimvalv{3.5cm}      
\def\trimvall{1.8cm}      
\def\trimvalr{2.8cm}      

\node[img, anchor=center, rotate=90, transform shape] (input) at (0, 0) {\includegraphics[height=\imh, trim=4.2cm 5cm 10cm 5cm, clip]{dense_bev_no_title_small.png}};

\node[img, anchor=center, rotate=90, transform shape] (out) at (\linewidth, 0) {\includegraphics[height=\imh, trim=4.2cm 5cm 10cm 5cm, clip]{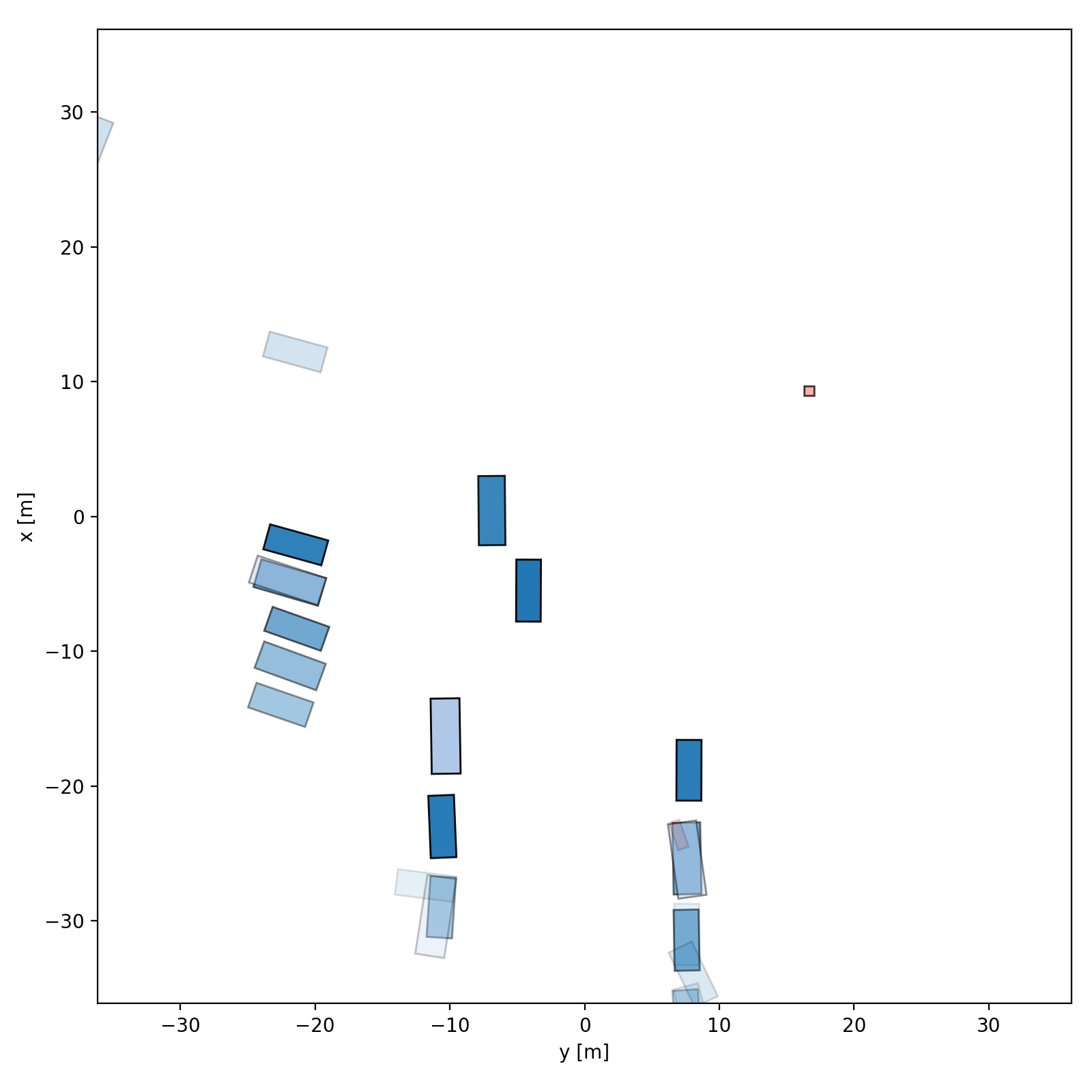}};

\coordinate (left_anchor) at (input.south);
\coordinate (right_anchor) at (out.north);

\node[trapezium, trapezium stretches=true, draw, rotate=90, transform shape] (in_mlp) at ($(left_anchor)!1/16!(right_anchor)$) {MLP};

\node[trapezium, trapezium stretches=true, draw, rotate=90, shape border rotate=180, transform shape] (out_mlp) at ($(left_anchor)!14/16!(right_anchor)$) {MLP};

\node[block, rotate=90, transform shape, minimum height=0.02\linewidth] (topk) at ($(left_anchor)!15/16!(right_anchor)$) {topK};

\node[img, rotate=90, transform shape] (d2d_attn_1) at ($(left_anchor)!3/16!(right_anchor)$)
  {\includegraphics[height=\imh, trim={\trimvall} {\trimvalv} {\trimvalr} {\trimvalv}, clip]{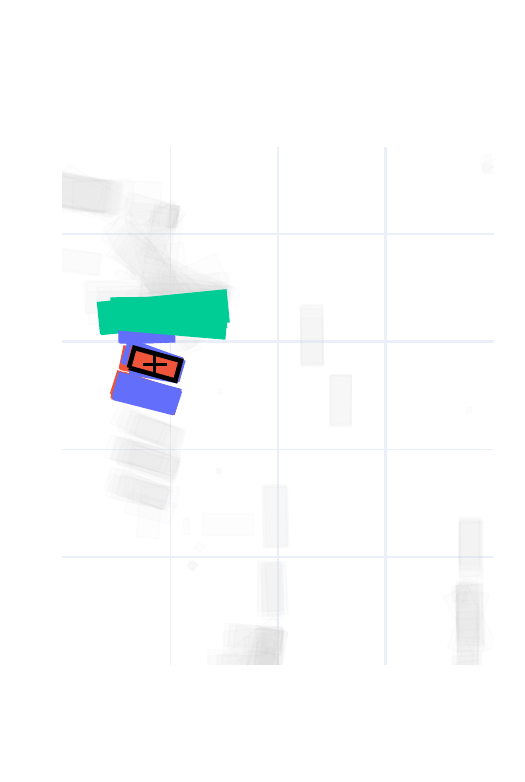}};

\node[img, rotate=90] (d2d_attn_2) at ($(left_anchor)!6/16!(right_anchor)$)
  {\includegraphics[height=\imh, trim={\trimvall} {\trimvalv} {\trimvalr} {\trimvalv}, clip]{gossipnet_0_0_3_attention_viz_no_axis.pdf}};

\node[img, rotate=90] (d2d_attn_3) at ($(left_anchor)!9/16!(right_anchor)$)
  {\includegraphics[height=\imh, trim={\trimvall} {\trimvalv} {\trimvalr} {\trimvalv}, clip]{gossipnet_0_0_3_attention_viz_no_axis.pdf}};

\node[img, rotate=90] (d2d_attn_4) at ($(left_anchor)!12/16!(right_anchor)$)
  {\includegraphics[height=\imh, trim={\trimvall} {\trimvalv} {\trimvalr} {\trimvalv}, clip]{gossipnet_0_0_3_attention_viz_no_axis.pdf}};

\node[block, anchor=center] (d2dlayer1) at ($(d2d_attn_1.east)+(0,18pt)$) {Gossip\\Block 1};
\node[block, anchor=center] (d2dlayer2) at ($(d2d_attn_2.east)+(0,18pt)$) {Gossip\\Block 2};
\node[block, anchor=center] (d2dlayer3) at ($(d2d_attn_3.east)+(0,18pt)$) {Gossip\\Block 3};
\node[block, anchor=center] (d2dlayer4) at ($(d2d_attn_4.east)+(0,18pt)$) {Gossip\\Block $L_\textrm{G}$};

\node[anchor=center] (dots) at ($(d2dlayer3)!0.5!(d2dlayer4)$) {\Large $\cdots$};

\node[] (d2dlbl) at ($(d2dlayer2.north)!0.5!(d2dlayer3.north)+ (0,10pt)$) {GossipNet3D: static local neighborhood};

\begin{pgfonlayer}{background}
  \node[fill=gray!15, draw, dashed, rounded corners,
        fit=(in_mlp) (topk) (d2dlayer1) (d2d_attn_1) (d2dlbl.west),
        inner sep=8pt] (background) {};
\end{pgfonlayer}

\node[] (apprlbl) at ($(background.west |- background.north)+(8pt, -8pt)$) {III};

\draw[->, thick] (input) -- (in_mlp);
\draw[->, thick] (in_mlp) |- (d2dlayer1);
\draw[->, thick] (d2dlayer1) -- (d2dlayer2);
\draw[->, thick] (d2dlayer2) -- (d2dlayer3);
\draw[->, thick] (d2dlayer4) -| (out_mlp);
\draw[->, thick] (out_mlp) -- (topk);
\draw[->, thick] (topk) -- (out);

\end{tikzpicture}

%% file: figures/attention_viz_legend.tex
\newcommand{\legendbox}[2]{%
  \tikz[baseline=-0.6ex]\node[fill=#1,minimum width=1.8em,minimum height=0.8em,inner sep=0pt] {};~#2%
}

\newcommand{\legendboxego}[1]{%
  \tikz[baseline=-0.6ex]{
    \node[draw, minimum width=1.8em, minimum height=0.8em, inner sep=0pt] (ego) {};
    \draw ($(ego.west)+(0.45em, 0)$) -- ($(ego.east)-(0.45em, 0)$);
    \draw ($(ego.north)-(0, 0.15em)$) -- ($(ego.south)+(0, 0.15em)$);
  }~#1%
}

\definecolor{clsCar}{HTML}{636EFA}
\definecolor{clsTruck}{HTML}{EF553B}
\definecolor{clsTrailer}{HTML}{00CC96}
\definecolor{clsBus}{HTML}{AB63FA}
\definecolor{clsConstr}{HTML}{FFA15A}
\definecolor{clsBicycle}{HTML}{19D3F3}
\definecolor{clsMotorcycle}{HTML}{FF6692}
\definecolor{clsPedestrian}{HTML}{B6E880}
\definecolor{clsCone}{HTML}{FF97FF}
\definecolor{clsBarrier}{HTML}{FECB52}
\definecolor{clsUnattended}{HTML}{E0E0E0}
\definecolor{clsEgo}{HTML}{000000}

\begin{center}
\begin{tabular}{llllll}
  \legendbox{clsCar}{car} &
  \legendbox{clsTruck}{truck} &
  \legendbox{clsTrailer}{trailer} &
  \legendbox{clsBus}{bus} &
  \legendbox{clsConstr}{construction vehicle} &
  \legendbox{clsBicycle}{bicycle} \\
  \legendbox{clsMotorcycle}{motorcycle} &
  \legendbox{clsPedestrian}{pedestrian} &
  \legendbox{clsCone}{traffic cone} &
  \legendbox{clsBarrier}{barrier} &
  \legendbox{clsUnattended}{unattended} &
  \legendboxego{target vehicle} \\
\end{tabular}
\end{center}

%% file: sections/related_work.tex
\section{Related Work}\label{sec:related_work}

Heuristic post-processing via \gls{nms} is widely used in 2D~\cite{wang2023internimage,lyu2022rtmdet} and 3D~\cite{yin2021center,lang2019pointpillars,wang2023dsvt} object detection due to its simplicity and strong empirical performance.
CenterPoint~\cite{yin2021center}, applies a distance-based variant (CircleNMS) that suppresses candidates based on Euclidean center distance in \gls{bev}.
However, \gls{nms} remains sensitive to fixed thresholds and may suppress valid detections or retain false positives in crowded scenes~\cite{bodla2017soft,hosang2017learning}.  
Soft-NMS~\cite{bodla2017soft} mitigates hard suppression by decaying scores according to overlap with higher-scoring boxes.
Though extendable to \gls{bev} or 3D using rotated-box \gls{iou}, it still depends on hand-tuned decay parameters and ignores contextual cues among detections.  

\label{sec:related_work:learned_nms}%
Learned NMS methods~\cite{hosang2017learning,ding2022end,hu2018relation,symeonidis2023neural} replace fixed rules with trainable modules that re-rank detections.
Early variants operate on pairwise local features~\cite{hosang2017learning}, while graph-based approaches~\cite{ding2022end} propagate information between spatially related detections.
Relation Networks~\cite{hu2018relation} integrate relation reasoning directly into the detector, requiring retraining and relying on 2D-specific geometric features.
Seq2Seq-NMS~\cite{symeonidis2023neural} refines 2D \gls{rois} via attention within \gls{iou}-based neighborhoods using image features from multiple encoder levels, tying it to a specific backbone and hindering 3D transfer.
Despite improved adaptability, these methods depend on hand-crafted neighborhood definitions, limiting global context modeling and robustness across data distributions.

Transformer-based set prediction detectors such as DETR~\cite{carion2020end} and its 3D extensions~\cite{wang2022detr3d,erabati2023li3detr} integrate suppression into training by predicting a fixed set of objects via one-to-one matching, replacing \gls{nms} with top-$K$ selection.
While elegant, their many learned queries and costly cross-attention~\cite{liu2023petrv2,yang2023bevformerv2} limit real-time use in automotive perception.  

Unlike heuristic suppression or feature-map-level set prediction, detection-level rescoring operates solely on dense proposals from a base detector, without accessing backbone features or raw point clouds.
This enables lightweight, detector-agnostic modules that reason over detection relations to adjust objectness scores before top-$K$ selection.
Our proposed \ApproachName{} follows this paradigm, providing an adaptive alternative to \gls{nms}.

%% file: sections/problem_formulation.tex
\section{Problem Formulation}
\label{sec:problem_formulation}

Let $\mathcal{D}_d = \{ \mathbf{d}_i \mid i = 1, 2, \dots, N \}$ denote the set of $N$ candidate detections generated by a base detector~\cite{yin2021center,lang2019pointpillars,wang2023dsvt}, where each $\mathbf{d}_i \in \mathbb{R}^{q_0}$ represents an object hypothesis with $q_0$ detection-level features predicted from a discrete reference location (anchor, center point, or similar).  
Each detection is encoded as
\[
\mathbf{d}_i = [x_i, y_i, z_i, l_i, w_i, h_i, \theta_i, c_{\mathrm{id},i}, s_i, v_{x,i}, v_{y,i}]^\top ,
\]
comprising the 3D bounding box center $(x_i, y_i, z_i)$, dimensions $(l_i, w_i, h_i)$, yaw $\theta_i$, class ID $c_{\mathrm{id},i}$, score $s_i$, and planar velocity $(v_{x,i}, v_{y,i})$.  
As each reference location contributes one detection, $\mathcal{D}_d$ contains redundancy and false positives.  

The scene includes $M \leq N$ ground-truth objects $\mathcal{G} = \{ \mathbf{g}_j \mid j = 1, 2, \dots, M \}$ in the same format as $\mathbf{d}_i$.  
A filtering function $f$ is applied to $\mathcal{D}_d$ to obtain a sparser set $\mathcal{D}_f = f(\mathcal{D}_d)$, which is evaluated against $\mathcal{G}$ using a task-specific similarity metric $S(\mathcal{D}_f, \mathcal{G})$ such as mAP or NDS~\cite{caesar2020nuscenes}.  
The filtering objective is
\begin{equation}
\mathcal{D}^*_f = \argmax_{\mathcal{D}_f \subseteq \mathcal{D}_d} S(\mathcal{D}_f, \mathcal{G})
\quad\text{s.t.}\quad \mathcal{D}_f = f(\mathcal{D}_d),\; |\mathcal{D}_f| \leq K ,
\label{eq:filtering_objective}
\end{equation}
where $K$ is the maximum number of retained detections.

In conventional post-processing, $f$ is rule-based, combining score thresholding with \gls{nms} to suppress overlaps and retain the top-$K$ predictions.  
Such heuristics rely on fixed thresholds and ignore variations in object density, occlusion, or detector uncertainty, often suppressing valid detections or retaining false positives~\cite{bodla2017soft,gahlert2020visibility}.  
These limitations motivate replacing $f$ with a learned function $f_\theta$, described in Section~\ref{sec:methodology}.

%% file: sections/methodology.tex
\section{Methodology}
\label{sec:methodology}
The proposed methodology consists of several key components to build the learned filtering function $f_{\boldsymbol{\theta}}$.
It consists of three stages: input embedding, context feature aggregation, and score refinement.
For context aggregation, we propose two interchangeable variants: the graph-based GossipNet3D and the transformer-based \ApproachName{}.
These share identical embedding, refinement, and supervision schemes.

\subsection{Input Embedding}
Each detection $\mathbf{d}_i \in \mathbb{R}^{q_0}$, as defined in Section~\ref{sec:problem_formulation}, is mapped to a p-dimensional latent vector
\begin{equation}
\mathbf{x}_i = \rho_{\boldsymbol{\theta}}(\mathbf{d}_i), \qquad \rho_{\boldsymbol{\theta}} : \mathbb{R}^{q_0} \rightarrow \mathbb{R}^p ,
\end{equation}
via a shared multilayer perceptron (MLP) after applying feature normalization, Fourier encodings~\cite{tancik2020fourier} to the coordinates, encoding the heading as a 2D direction vector, and mapping the discrete class ID to a one-hot representation.
The resulting set of embedded detections is
\begin{equation}
\mathcal{X} = \{ \mathbf{x}_i \mid i = 1, 2, \dots, N \}, \quad \mathbf{x}_i \in \mathbb{R}^p .
\end{equation}

\subsection{Context Feature Aggregation}
For the sake of brevity, we define a shorthand for the definition of input and output sets
\begin{equation}
    \left\lbrace\mathbb{R}^p\right\rbrace^N = \left\lbrace\mathbb{X} \subset \mathbb{R}^p: \left\lvert\mathbb{X}\right\rvert = N\right\rbrace
\end{equation}
defining the set of sets $\mathbb{X}$ of vectors $x \in \mathbb{R}^p$ that have cardinality $N$.

Both aggregation modules map the embedded set $\mathcal{X}$ to context-enriched embeddings $\mathcal{Z} = \{\mathbf{z}_i\}_{i=1}^{N}$ using different mechanisms:

\vspace{1em}

\textbf{A) GossipNet3D.}
We adapt GossipNet~\cite{hosang2017learning} from 2D to 3D detection.
The embedded detection set $\mathcal{X}$ is processed by a graph-based encoder
\begin{equation}
\mathcal{Z} = \gamma_{\boldsymbol{\theta}}(\mathcal{X}), \quad \gamma_{\boldsymbol{\theta}} :  \left\lbrace\mathbb{R}^p\right\rbrace^N \rightarrow  \left\lbrace\mathbb{R}^p\right\rbrace^N ,
\end{equation}
which consists of $L_\textrm{G}$ stacked GossipBlocks.  
Each block updates the embeddings by exchanging information within local neighborhoods as follows.

For each detection $\mathbf{x}_i \in \mathcal{X}$, we define a local neighborhood
\begin{equation}
\mathcal{N}_i = \{\, \mathbf{x}_j \mid \| \mathbf{c}_i - \mathbf{c}_j \| \leq \epsilon, \; \mathbf{x}_j \in \mathcal{X} \,\},
\end{equation}
where $\mathbf{c}_i$ and $\mathbf{c}_j$ denote the 3D box centers and $\epsilon$ is a fixed radius threshold.

For each pair $(\mathbf{x}_i, \mathbf{x}_j)$, with $\mathbf{x}_j \in \mathcal{N}_i$, we form a pair feature vector by concatenating both embeddings with hand-crafted geometric relations:
\[
\mathbf{u}_{ij} = \text{concat}\big( \mathbf{x}_i,\, \mathbf{x}_j,\, \Delta \mathbf{c}_{ij},\, \Delta \mathbf{s}_{ij},\, \cos(\Delta \theta_{ij}),\, \|\mathbf{c}_i - \mathbf{c}_j\| \big),
\]
where $\Delta \mathbf{c}_{ij}$ denotes the normalized center offset between the boxes,  
$\Delta \mathbf{s}_{ij}$ are the normalized differences in box dimensions,  
$\Delta \theta_{ij}$ is the difference in yaw angle,  
and $\|\mathbf{c}_i - \mathbf{c}_j\|$ is the Euclidean distance between centers.  

An MLP maps each pair feature to a message:
\[
\mathbf{m}_{ij} = \mu_{\boldsymbol{\theta}}(\mathbf{u}_{ij}), \quad \mu_{\boldsymbol{\theta}} : \mathbb{R}^{2p+8} \rightarrow \mathbb{R}^p,
\]
which is aggregated for each target detection via max pooling:
\[
\mathbf{m}_i = 
\left( \max_{j} \mathbf{m}_{ij}^{\left[1\right]}, \max_{j} \mathbf{m}_{ij}^{\left[2\right]}, \dots, \max_{j} \mathbf{m}_{ij}^{\left[p\right]} \right)^\text{T}
\]
with $\mathbf{x}^{[k]}$ denoting the $k$-th element of vector $\mathbf{x}$.

The aggregated message is added to the original embedding to form an updated representation:
\[
\mathbf{x}'_i = \mathbf{x}_i + \mathbf{m}_i.
\]
Applying these operations to all detections defines one GossipBlock.
GossipNet3D produces a context-enriched set $\mathcal{Z} = \gamma_{\boldsymbol{\theta}}(\mathcal{X}) = \{ \mathbf{z}_i \mid i = 1, 2, \dots, N \}, \quad \mathbf{z}_i \in \mathbb{R}^p$.
\vspace{1em}

\textbf{B) \ApproachName.}
The embedded detection set $\mathcal{X}$ is processed by a transformer encoder
\begin{equation}
\mathcal{Z} = \tau_{\boldsymbol{\theta}}(\mathcal{X}), \quad \tau_{\boldsymbol{\theta}} : \left\lbrace\mathbb{R}^p\right\rbrace^N \rightarrow \left\lbrace\mathbb{R}^p\right\rbrace^N,
\end{equation}
consisting of $L$ layers of multi-head self-attention and feed-forward sublayers with residual connections.
Self-attention enables each detection to attend to all others in the frame, capturing global spatial relations and redundancy in the candidate set, which we term detection-to-detection (D2D) attention.
A learnable temperature parameter in the attention softmax allows the model to balance between focused and distributed interactions.
The resulting set $\mathcal{Z} = \{ \mathbf{z}_i \mid i = 1, 2, \dots, N \}$ with $\mathbf{z}_i \in \mathbb{R}^p$ contains context-enriched embeddings for all detections.

\subsection{Score Refinement Head}
For each context-enriched embedding $\mathbf{z}_i \in \mathbb{R}^p$ from $\mathcal{Z}$, a score MLP head
\begin{equation}
\Delta \ell_i = \psi_{\boldsymbol{\theta}}(\mathbf{z}_i), \quad \psi_{\boldsymbol{\theta}} : \mathbb{R}^p \rightarrow \mathbb{R}
\end{equation}
predicts a residual adjustment in logit space.
Let $\ell_i = \sigma^{-1}(s_i)$ be the logit of the original detector score $s_i$, derived from the inverse sigmoid function $\sigma^{-1}$.
The refined logit is computed as
\begin{equation}
\hat{\ell}_i = \ell_i + \Delta \ell_i .
\end{equation}
This residual formulation preserves the base detector's prior confidence while enabling context-driven adjustments derived from detection-to-detection attention.  
The score head is implemented as a multilayer perceptron whose last layer weights are initialized to zero.
Refined scores are supervised by the binary cross-entropy loss, as described in Section~\ref{sec:supervision}.

\subsection{Inference}
At inference time, the refined logits $\hat{\ell}_i$ are converted to objectness scores via the sigmoid function
\[
\hat{s}_i = \sigma(\hat{\ell}_i) .
\]
The learned filtering function is then applied as
\begin{equation}
\mathcal{D}_f = f_{\boldsymbol{\theta}}(\mathcal{D}_d) = \text{topK}\left(\mathcal{D}_d, \hat{\mathbf{s}}\right),
\end{equation}
where $\text{topK}(\cdot, \hat{\mathbf{s}})$ returns the $K$ detections with the highest refined scores $\hat{s}_i$.
This selection step replaces heuristic NMS while remaining detector-agnostic and operating solely on detection-level inputs.

\subsection{Supervision}
\label{sec:supervision}
Training follows a one-to-one assignment between detections in $\mathcal{D}_d$ and ground-truth objects in $\mathcal{G}$.
Instead of Hungarian matching with hand-tuned cost weights, we adopt a metric-aware greedy matching procedure aligned with the nuScenes evaluation protocol~\cite{caesar2020nuscenes}.
Thus, we apply the thresholds defined for the \gls{map} calculation directly during training.

For each semantic class, candidates are first sorted in descending order by their refined score $\hat{s}_i$.
Each candidate is matched to the closest unmatched ground-truth object of the same class if the center distance is below a threshold $\tau_d$.
Matching proceeds sequentially with thresholds
\[
\tau_d \in \{ \qty{0.5}{\meter}, \qty{1}{\meter}, \qty{2}{\meter}, \qty{4}{\meter} \}
\]
until no further matches can be made.
A second, class-agnostic matching stage preserves correctly localized but misclassified detections, prioritizing safety over marginal metric degradation.

Matched detections form the positive set, and unmatched detections form the negative set for the binary cross-entropy loss.

%% file: sections/evaluation.tex
\section{Experimental Evaluation}\label{sec:evaluation}

\subsection{Experimental Setup.}
\label{sec:experiments}
\textbf{Dataset.}
We focus on CenterPoint \cite{yin2021center} on nuScenes \cite{caesar2020nuscenes} as a representative baseline for high-performance 3D detection.
However, our module's reliance on universal detection features makes it architecturally compatible with any 3D detector.

\textbf{Detection extraction.}
For each frame, we extract the dense set $\mathcal{D}_d$ of pre-NMS detections from the base detector, filtered by a confidence threshold of $0.1$.
These outputs are stored and reused across all post-processing methods to ensure comparability.

\textbf{Baselines.}
We use CenterPoint~\cite{yin2021center} as base detector and compare \ApproachName{} to four post-processing methods:
(1) CircleNMS, the default distance-based suppression in CenterPoint;
(2) Soft-CircleNMS, an adaptation of Soft-NMS~\cite{bodla2017soft} with Gaussian score decay;
(3) the graph convolutional duplicate removal of~\cite{ding2022end} (GCN);
and (4) GossipNet3D, our 3D adaptation of the learned NMS in~\cite{hosang2017learning}.

\textbf{Evaluation metrics.}
We follow the official nuScenes evaluation protocol, reporting \gls{map} and \gls{nds}, as well as per-class AP.
As the work at hand does not consider object attributes, the reported \gls{nds} does not capture the up to $10$ percentage points for correct attributes.
In addition, we analyze true positive quality metrics, namely translation, scale, orientation, and velocity errors.
All metrics are computed on the validation set using the official evaluation scripts.

\textbf{Implementation details.}
All experiments are run with a batch size of 32 on a workstation with an NVIDIA RTX~3090 GPU and an AMD Ryzen~9~5900X CPU.
The transformer encoder in \ApproachName{} has $L=6$ layers, $p=64$ feature channels, and $h=4$ attention heads.
GossipNet3D uses $L_\textrm{G} = 4$ blocks, $p=64$ channels, and a neighborhood radius of $\epsilon = \qty{5}{\meter}$.
The input MLP consists of two layers with ReLU activation and batch normalization.
Location coordinates are augmented with Fourier features~\cite{tancik2020fourier} using $F=10$ frequencies.
The score head consists of two layers with ReLU activation and instance normalization, with the last layer weights initialized to zero.
Training uses the AdamW optimizer with a maximum learning rate of $5.5 \times 10^{-4}$ and weight decay of $1 \times 10^{-2}$ with a cosine annealing learning rate schedule and constant warmup over \qty{100}{epochs}.
A maximum of $K=300$ detections per frame is retained unless stated otherwise.

\subsection{Comparison with Baselines}

\begin{table*}
    \vspace{4pt}
    \centering
    \caption{Comparison of post-processing methods on CenterPoint detections (nuScenes val).}
    \label{tab:centerpoint_baselines}
    \begin{tabular}{lcccccccc}
        \toprule
        Method & mAP [\%] & NDS [\%] & $T_\text{CPU}$~/~$T_\text{GPU}$ [ms] & $M_\text{GPU}$ [MB] & $e_\text{t}$ $\downarrow$ & $e_\text{s}$ $\downarrow$ & $e_\text{o}$ $\downarrow$ & $e_\text{v}$ $\downarrow$ \\
        \midrule
        CircleNMS
        & 56.22 & 56.53 & \textbf{4.74}~/~14.77\textsuperscript{2} & - & 29.05 & 25.35  &  32.88 &  28.52 \\
        
        Soft-CircleNMS 
        & 53.71 & 55.56 & \underline{6.74}~/~26.84\textsuperscript{2} & - &  \textbf{28.38} &  \textbf{25.14} &  \textbf{31.46} &   27.98 \\
        
        GCN 
        & 53.73 & 55.36 & 346.27~/~\textbf{3.42} & \textbf{44.58} &  29.40 & 25.42  &  \underline{32.14} & 28.03 \\
        
        GossipNet3D 
        & \textbf{58.39} & \textbf{57.78} & 180.66~/~7.75 & 263.71 & 29.07 &  \underline{25.30} & 32.51  &  \textbf{27.25} \\

        \textbf{\ApproachName{} (Ours)}
        & \underline{57.50} & \underline{57.34} & 9.35~/~\underline{6.66} & \underline{51.57} & \underline{28.97} & 25.31  & 32.59  &  \underline{27.38} \\
        \bottomrule
    \end{tabular}
\end{table*}
We first evaluate \ApproachName{} on CenterPoint detections using the four baseline post-processing methods described in Section~\ref{sec:experiments}.
Table~\ref{tab:centerpoint_baselines} reports mAP, NDS, and the true positive quality metrics for translation ($e_\text{t}$), scale ($e_\text{s}$), orientation ($e_\text{o}$), and velocity ($e_\text{v}$) from the nuScenes devkit.

\ApproachName{} surpasses the default CircleNMS by a clear margin in both mAP and NDS, while also improving translation, scale, orientation, and velocity errors.
Note that these scores are computed over all \gls{tps}, inherently increasing difficulty when producing more \gls{tps}.
Soft-CircleNMS performs worse than CircleNMS in our setting, consistent with the observation that in the \gls{bev}, overlaps correspond to actual spatial collisions, a scenario that is rare and thus not represented in the annotations.
As a result, the soft decay of overlapping scores is less effective than hard suppression, in contrast to 2D image detection where overlaps frequently occur due to depth ambiguity.
The GCN baseline also performs worse than CircleNMS, indicating that its graph-construction and message-passing approach does not translate effectively to dense 3D candidate sets from CenterPoint.
GossipNet3D achieves the highest mAP among all methods, outperforming \ApproachName{} by roughly 1\,pp.
Training GossipNet3D is slower and more memory-intensive, while \ApproachName{} trains faster and with lower peak memory usage.
At inference, \ApproachName{} remains smaller in model size, is comparatively fast on GPU, and is significantly faster on CPU, making it more practical for deployment in resource-constrained settings.
It remains an open question why the neighborhood-constrained architecture of GossipNet3D yields better accuracy than the fully unconstrained transformer attention used in \ApproachName{}.
The investigation of the resulting neighborhoods in Section~\ref{sec:qualitative} aims to give insights into possible reasons for this phenomenon.

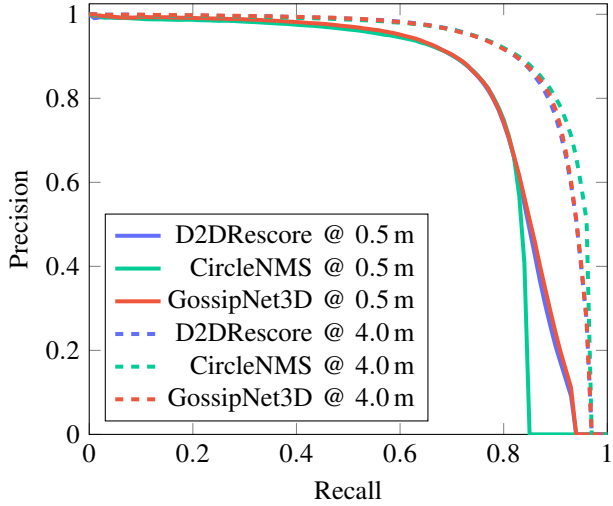
\begin{figure}
    \centering
    \input{figures/pr_curve_car_D2DRescore_CircleNMS_GossipNet3D.tex}   
    \caption{Precision-recall curves of \ApproachName{} (blue), CircleNMS (green), and GossipNet3D (red) for car detections at \qty{0.5}{\meter} (solid) and \qty{4.0}{\meter} (dashed) assignment thresholds on the nuScenes validation split.}
    \label{fig:pr_curve}
\end{figure}
As shown in Table~\ref{tab:centerpoint_baselines}, \ApproachName{} achieves improvements over CircleNMS in all four true positive metrics while increasing the number of retained true positives.  
This is corroborated by the maximum recall values of the precision-recall-curve (Figure~\ref{fig:pr_curve}), which are higher for all classes and center-distance thresholds.  
At the \qty{0.5}{\meter} threshold, recall increases from $0.84$ to $0.92$ for car and from $0.61$ to $0.81$ for truck.  
These results indicate that the additional detections retained by \ApproachName{} are more numerous and also exhibit improved spatial accuracy compared to those retained by heuristic \gls{nms}.
Figure~\ref{fig:pr_curve} reveals that the improvements in precision are most pronounced at high recall levels, which typically correspond to more challenging detection scenarios.

\begin{table*}
    \vspace{4pt}
    \centering
    \caption{Per-class AP [\%] on CenterPoint detections (nuScenes val). 
    C.V.~=~Construction Vehicle, Ped.~=~Pedestrian, Motor.~=~Motorcycle, 
    T.C.~=~Traffic Cone.}
    \label{tab:per_class_ap}
    \begin{tabular}{lcccccccccc}
        \toprule
        Method & Car & Truck & Bus & Trailer & C.V. & Ped. & Motor. & Bicycle & T.C. & Barrier \\
        \midrule
        CircleNMS &
        84.90 & 53.32 & \underline{66.52} & 32.78 & 15.48 & 83.64 & 54.66 & 36.93 & 66.43 & \textbf{67.54} \\
        
        Soft-CircleNMS &
        78.52 & 47.79 & 62.85 & 31.72 & 14.91 & 81.22 & 54.56 & 37.82 & 64.39 & 63.29 \\
        
        GCN &
        82.89 & 49.28 & 63.21 & 25.72 & 13.10 & 82.25 & 54.24 & 37.83 & 66.35 & 62.38 \\
        
        GossipNet3D &
        \textbf{85.04} & \textbf{53.75} & \textbf{66.90} & \textbf{34.24} & \textbf{18.63} & \textbf{85.26} & \textbf{61.33} & \textbf{43.41} & \textbf{68.67} & 66.70 \\
        
        \textbf{\ApproachName{} (Ours)} &
        \underline{84.91} & \underline{53.62} & 66.12 & \underline{34.03} & \underline{17.29} & \underline{84.47} & \underline{57.59} & \underline{40.49} & \underline{68.58} & \underline{66.79} \\
        \bottomrule
    \end{tabular}
\end{table*}

As shown in Table~\ref{tab:per_class_ap}, \ApproachName{} improves AP in $8$ out of $10$ classes compared to CircleNMS.
The largest gains are observed for Construction Vehicle (\qty{+1.81}{pp}), Motorcycle (\qty{+2.93}{pp}), Bicycle (\qty{+3.56}{pp}), and Traffic Cone (\qty{+2.15}{pp}).
Except for Traffic Cone, these categories are the rarest in the nuScenes dataset, indicating that the learned rescoring appears particularly beneficial for rare classes.
For Traffic Cone, which is the smallest object category, the improvement suggests that \ApproachName{} can better retain small objects that might otherwise be suppressed by hard NMS.

\subsection{Qualitative Comparison of Neighborhoods}
\label{sec:qualitative}

The attention maps of \ApproachName{} and GossipNet3d shown in Figure~\ref{fig:overview}, give insights into how both methods leverage inter-detection relations for score refinement. 
GossipNet3Ds neighborhood map shows a clear local focus on nearby detections, while \ApproachName{} exhibits a broader and more blurred context awareness.
While GossipNet3D receives features from each neighbor through maxpooling, \ApproachName{} receives a weighted sum of features from all neighbors.
This may lead to more diffuse information propagation.

\subsection{Ablation Studies}
\begin{table}
    \centering
    \caption{Ablation of matching strategy and retained detections $K$ (nuScenes val) grouped by post-processing method.}
    \label{tab:ablation}
    \begin{tabular}{l c c c}
        \toprule
        Matching & $K$ & mAP [\%] $\uparrow$ & NDS [\%] $\uparrow$ \\
        \midrule
        \multicolumn{4}{c}{\cellcolor{gray!20}\textbf{\ApproachName}} \\
        Hungarian    & 300 & 56.29 & 56.66 \\
        Metric-aware & 300 & \textbf{57.39} & \underline{57.27} \\
        Metric-aware & 150 & \underline{57.23} & \textbf{57.34} \\
        Metric-aware & 100 & 56.78 & 57.21 \\
        \midrule
        \multicolumn{4}{c}{\cellcolor{gray!20}\textbf{GossipNet3D}} \\
        Hungarian    & 300 & 57.23 & 57.00 \\
        Metric-aware & 300 & 58.39 & 57.78 \\
        \bottomrule
    \end{tabular}
\end{table}
We conduct ablation studies to quantify the effect of individual design choices in \ApproachName{}.

\textbf{Matching strategy.}
We compare the proposed metric-aware matching (Sec.~\ref{sec:supervision}) to Hungarian matching with hand-tuned cost weights, keeping all other settings identical.
Table~\ref{tab:ablation} shows that metric-aware matching yields higher mAP and NDS and converges more stably, likely due to its closer alignment with the nuScenes evaluation procedure.

\textbf{Number of retained detections $K$.}
We vary $K$ in the top-$K$ selection stage to assess the ranking quality of the refined scores in Table~\ref{tab:ablation}.
Reducing $K$ can benefit subsequent modules such as multi-object tracking, which must process every retained detection.
Conventional post-processing often retains up to $500$ detections per frame, which may overwhelm such downstream tasks.
Our results show that with \ApproachName{}, $K$ can be reduced to as few as $150$ detections with only a negligible drop in mAP, indicating that the model effectively ranks the detections by quality.

\textbf{Runtime and memory breakdown.}
We measure the average inference time $T$ on CPU and GPU, and peak allocated GPU memory $M_\text{GPU}$ over the whole validation set, shown in Table~\ref{tab:centerpoint_baselines}.
On the GPU, \ApproachName{} runs at around $\qty{150}{\hertz}$ compared to GossipNet3D, which runs at $\qty{129}{\hertz}$.
A noticable difference is observed when executing the models on the CPU, where \ApproachName{} runs at $\qty{107}{\hertz}$ compared to only $\qty{5.5}{\hertz}$ for GossipNet3D.
During inference \ApproachName{} requires only $\qty{19.56}{\percent}$ of the GPU memory compared to GossipNet3D.
CircleNMS, implemented in C++, shows the lowest latency on CPU, but is slower on GPU due to its sequential processing of detections.

%% file: figures/pr_curve_car_D2DRescore_CircleNMS_GossipNet3D.tex
\pgfplotstableread{data0 D2DRescore_@_0.5m CircleNMS_@_0.5m GossipNet3D_@_0.5m D2DRescore_@_4.0m CircleNMS_@_4.0m GossipNet3D_@_4.0m
0.0 1.0 1.0 1.0 1.0 1.0 1.0
0.01 0.9906617031664109 0.9962436344143661 0.9962436344143661 1.0 0.9981182829956247 1.0
0.02 0.9925153894395335 0.995308951897753 0.9971800797855409 1.0 0.9990582561666764 1.0
0.03 0.9931348252301591 0.9918967262529036 0.9956203188632444 0.9993719736472696 0.9981182844758754 0.9987447356337049
0.04 0.9929798941075644 0.9925153898985885 0.9943760190974076 0.999058256225201 0.998588049185776 0.998588049185776
0.05 0.9917730866285939 0.9917730866285939 0.994003335537177 0.9992464630724756 0.9977427909981603 0.9988701203117556
0.06 0.9918967267092474 0.9912788341478039 0.9922059619426644 0.9993719736831725 0.9981182845831794 0.9987447357053756
0.07 0.9917201075793366 0.9911906268207306 0.9922501543233455 0.9994616434406814 0.9983866674184007 0.9989238662250396
0.08 0.9915876842567374 0.9904304878699827 0.9920513203089073 0.9990582562544427 0.998353111655389 0.998823097403222
0.09 0.9908673334451898 0.9898400744290652 0.9925153901922461 0.9991628068533006 0.9979096421831221 0.998953727549658
0.1 0.9910318927653828 0.9891837457017403 0.9932588056354384 0.9986820552309029 0.9973675798462034 0.9990582562602893
0.11 0.990325422189463 0.9883124735734482 0.9935294147326402 0.9984598877682662 0.9972652985025571 0.9989727311423968
0.12 0.9908159198724628 0.9880474871395101 0.9931348257581419 0.9984314119123182 0.9974926188886025 0.9985880492441721
0.13 0.9909464416908116 0.9878233787656959 0.9929441478552684 0.9982627804978527 0.9971079835413663 0.9984073182198763
0.14 0.99079388713785 0.9878941388965892 0.9930462876512468 0.9978500460888762 0.9971800800847291 0.9985209129457905
0.15 0.9907850742755384 0.9873424789644725 0.9930108767960003 0.9976176892747207 0.9971175958130695 0.9984940609470131
0.16 0.9903149168810587 0.9870898407329972 0.992167297168454 0.9975317000203764 0.9967116389143729 0.9981182846255184
0.17 0.9905528867827231 0.9867590396572323 0.9919694707673385 0.9975661859082618 0.9966840972533944 0.9981182846233358
0.18 0.9897374656925871 0.986873032219877 0.9917936913052581 0.9974926188916118 0.9965555896445186 0.9980139525021121
0.19 0.98968886897274 0.9869750478955918 0.991538905772193 0.9974268049690443 0.9966377147489696 0.997723036041191
0.2 0.989368250215622 0.9868832328372352 0.991402351252001 0.9975551501650108 0.9964307854075047 0.9977427910340885
0.21 0.988727173633559 0.9860140231515696 0.9908379536414941 0.9972247164951903 0.99615454438863 0.9974926188962927
0.22 0.9886473965905835 0.9857244945457827 0.9907458190463639 0.9970948762108249 0.9960735643258352 0.9974357790825865
0.23 0.9880941678354299 0.9855398989639325 0.9904204372602661 0.9967319967723451 0.9959996375155646 0.9972208349084191
0.24 0.9875109763615515 0.9849133328804255 0.9900453558343192 0.9967116389147765 0.9956981913574934 0.9971019759785615
0.25 0.9869015944939971 0.984410667541377 0.9897744024076348 0.9966180034911478 0.9954210220080357 0.9969176987957638
0.26 0.9864103029827626 0.9846496332217808 0.9894534293493356 0.9959558552685019 0.9952371261648669 0.9967476572265832
0.27 0.9860916121936746 0.9845324786838332 0.9888832192413789 0.9959665109090766 0.9951360549597714 0.9964515839462188
0.28 0.9855996135097728 0.9840323751848121 0.9888807806605198 0.9960432002053385 0.9949089096751419 0.9964441558017181
0.29 0.985520682025139 0.983127184581019 0.9885607568291238 0.9959856181491015 0.9946332054514708 0.9963081634240318
0.3 0.9856301370661875 0.9825264418712696 0.9881395189943872 0.9959942166488764 0.9943138859882917 0.9960565601924013
0.31 0.9853781281850524 0.9819064833764245 0.987330622116399 0.9960022605367063 0.9940153533341441 0.995881615832574
0.32 0.9848561859776245 0.9813827203779226 0.9870898407434772 0.9959513599884449 0.9940266201876925 0.9957176613006916
0.33 0.984587857361553 0.9806164528127088 0.9864188549055951 0.9957902377868907 0.9935858102196757 0.9956203192139818
0.34 0.9839058310970511 0.9797898799313134 0.9857342904922759 0.9955287205153259 0.9932806877002883 0.9954737693864052
0.35000000000000003 0.9830553797822819 0.9791667216039941 0.985351134887019 0.9951222257652387 0.9929931723536362 0.9952822728716209
0.36 0.9824051265788601 0.9783282304627647 0.9846340110079385 0.9947386187432066 0.992515390223795 0.9949977802616027
0.37 0.981545303275904 0.9776337726997903 0.984055283038006 0.9944264034431081 0.9923146581194966 0.9946784005389468
0.38 0.9806366927757645 0.9771662269469356 0.982980043238188 0.994081772150916 0.9920757343191815 0.9941308012558626
0.39 0.9801476089161122 0.9761699942855574 0.9824284541196113 0.9937550350004765 0.9918491693551201 0.994041546754363
0.4 0.9793668817610445 0.9750461852288363 0.9815416215552862 0.9936309311930895 0.9914486780467503 0.9938170985307957
0.41000000000000003 0.9785813568609332 0.9742413864195669 0.9808764623890506 0.9932860249261357 0.9911131892533207 0.9937398987295339
0.42 0.9776627612313619 0.9733485494154738 0.9795972748268804 0.9928692587497349 0.9907498245569388 0.9934005349126599
0.43 0.9764957621623098 0.9727891040283633 0.9788425546938455 0.992688178216972 0.9906186847091153 0.993077171250523
0.44 0.9757106211026316 0.9718508327287256 0.9784102570954117 0.9922622074296323 0.9904094730629276 0.9927264746105997
0.45 0.9749615539262911 0.9706795551803347 0.9773968269481187 0.991937947123895 0.9900864223352677 0.9925979369839077
0.46 0.9738184163422845 0.9692537235087624 0.9763903483679278 0.991627983341378 0.9895769033883361 0.9923539254030512
0.47000000000000003 0.9730301574291523 0.968118002990718 0.9751997030160158 0.9914496637706978 0.98908955727519 0.9918835722287045
0.48 0.9716822426958123 0.966701648276334 0.9738006030745135 0.9910859007724535 0.9885845812598381 0.9912402416261158
0.49 0.9700668866959731 0.9653469924835366 0.9722441156190594 0.9906617111471661 0.9877627353999258 0.9905106960452643
0.5 0.9685565598361334 0.9639099383842005 0.9712878357743723 0.9899960805504636 0.9873792371654487 0.989922176806299
0.51 0.9671098847996349 0.9627045334245884 0.9699185362384756 0.9893573951282711 0.9869030346642341 0.9896107430760634
0.52 0.9657229207475938 0.9613472487776673 0.9684354960488074 0.9885314360471767 0.9862692177275441 0.9890631454504708
0.53 0.9640613302930249 0.9596515284721788 0.9665140085863378 0.9882240413027277 0.9857983385079062 0.9884672683119113
0.54 0.9627578297557277 0.9575758669786671 0.9650609058759672 0.9874854427748933 0.9852436449611658 0.9875535353105983
0.55 0.9610299555355398 0.9555841616156431 0.9631875306335376 0.9872088360515073 0.9846764761930944 0.9866746258122465
0.56 0.9596176012953377 0.9533347369250535 0.9618237559177718 0.9865161434299424 0.9837390727217933 0.9858613053787266
0.5700000000000001 0.9571969172808695 0.9512341074195131 0.959292469477355 0.985720112240037 0.9830279833395305 0.9849494291004898
0.58 0.9548416291253475 0.9499474154498165 0.9570396759256262 0.9847951052394003 0.981778111649665 0.9839447032543331
0.59 0.9528381880871909 0.9475586460450797 0.9546693746990339 0.983532043528183 0.9807272682785947 0.9826364740133613
0.6 0.9504551318736638 0.9451205388087841 0.9517057008136447 0.9823748024469378 0.9800454520229133 0.9817989992506789
0.61 0.946912853666038 0.9425545680538625 0.9483560548050685 0.9813176118175706 0.9790905319413442 0.9804849844954193
0.62 0.9432122262251249 0.9398965294208212 0.945408539025789 0.9801506050449427 0.9780227362036413 0.9788960835970788
0.63 0.9394452067637111 0.9372319632303782 0.9421202483519596 0.9785072745832843 0.9765057162102018 0.9774482868140916
0.64 0.9359534961053565 0.9328158643462782 0.9377633249348596 0.9768081046130992 0.9750685900949134 0.9762743683756511
0.65 0.9313343210866545 0.9296518047429054 0.9326443581511406 0.9750565258142755 0.9735972292611765 0.9747532933089469
0.66 0.9271439308469953 0.9251104185184376 0.928127132512546 0.973553492852956 0.971769902213811 0.9726608798812706
0.67 0.9214868950708314 0.9203176108356248 0.9227070649608862 0.9716213633687737 0.9698183726392153 0.9703744640406202
0.68 0.9160618719145097 0.9151321958664497 0.9159455591222856 0.9692582532413013 0.9680616137055842 0.9680875961892933
0.6900000000000001 0.9093612126988239 0.9096324231584416 0.9104243784869093 0.9668724327074892 0.9658263286050679 0.965928287881982
0.7000000000000001 0.9025940088092477 0.9031648386880432 0.9042866432324863 0.9641902822947384 0.9634147290721758 0.9628900627812326
0.71 0.895154751904179 0.8955804937271354 0.8964331934444167 0.961352439959184 0.9603718315570344 0.9595398859458739
0.72 0.8864722032121752 0.8871105313307602 0.8886793611884811 0.9579361472209635 0.9573118336700319 0.9562092701133876
0.73 0.8764582308477946 0.8776899539918489 0.8784266654479209 0.9539311169124782 0.9543072783707587 0.9535552518821797
0.74 0.8648286000112018 0.8666617208424455 0.8653815372864465 0.9498364421144684 0.9505495011075081 0.9496296253242533
0.75 0.8520461980693154 0.854553656284238 0.8531059513407432 0.9456822112155632 0.94671743462563 0.9454349931320067
0.76 0.8371150170498461 0.8397481297895083 0.8398705964262778 0.9413862000160232 0.9419140964091572 0.9409247758292462
0.77 0.8192690486038903 0.8241957287483106 0.8233317764258259 0.9367232457295743 0.9372176255976536 0.9356931746530712
0.78 0.7980615918709312 0.8028156879514521 0.8000523653774152 0.9313846390651092 0.9321190613104811 0.9300655913180059
0.79 0.7726517162560295 0.7784939408605064 0.7750669241077712 0.9249309733557219 0.9263827605658455 0.9237280158670211
0.8 0.7405344278584017 0.7491763974572878 0.7443012397114244 0.9185451995042326 0.9199790137231034 0.9169771200778243
0.81 0.7042049058242625 0.7103875723816014 0.706218975912169 0.9111824253122075 0.9136432041654743 0.9107961086865718
0.8200000000000001 0.6589266414894799 0.6588068752248245 0.662459261915756 0.9022741029818526 0.906618536296114 0.9039617792325502
0.8300000000000001 0.6061832614921409 0.5683690445805821 0.6164967318672816 0.8930941021321767 0.8990028728753052 0.8940190302657942
0.84 0.5476475163376014 0.40695307356178523 0.564257792012591 0.8826818431382661 0.8894303477018705 0.8844690698640114
0.85 0.48727703260951494 0.0 0.5111879822360582 0.8693637193523248 0.8797624518043793 0.8722397033768893
0.86 0.4277060480429027 0.0 0.4597420879708841 0.8529092262757202 0.8666172220125151 0.8607467821741634
0.87 0.3653752558960706 0.0 0.40179399112956315 0.8355888860129395 0.8544011348592438 0.845540490091329
0.88 0.30964390364258715 0.0 0.3478155901868161 0.8146526989423203 0.8387926864173747 0.8281963150228121
0.89 0.25862972064720097 0.0 0.2944016670139169 0.7901987989292646 0.8225545007958496 0.8053782678315352
0.9 0.21064347730609298 0.0 0.24305920795972194 0.757352019621117 0.8020655345927757 0.774416940941095
0.91 0.17037618942837743 0.0 0.19545708566423195 0.7149167922203218 0.7789976071188884 0.7318258086660252
0.92 0.13096490119582052 0.0 0.15357336983338543 0.6622406517222021 0.747959798623008 0.6730550464534769
0.93 0.09114422872240206 0.0 0.11173599928727251 0.595025522896445 0.7116682600935054 0.6024154752782088
0.9400000000000001 0.0 0.0 0.0 0.5090685104754206 0.6602505752832135 0.5120852539666688
0.9500000000000001 0.0 0.0 0.0 0.3932519294452563 0.5896642574514247 0.4021524650029243
0.96 0.0 0.0 0.0 0.2728881602484959 0.5089467226435076 0.2711791330624106
0.97 0.0 0.0 0.0 0.0 0.0 0.0
0.98 0.0 0.0 0.0 0.0 0.0 0.0
0.99 0.0 0.0 0.0 0.0 0.0 0.0
1.0 0.0 0.0 0.0 0.0 0.0 0.0
}\dataZ

\begin{tikzpicture}

\definecolor{00CC96}{HTML}{00CC96}
\definecolor{636EFA}{HTML}{636EFA}
\definecolor{EF553B}{HTML}{EF553B}

\begin{axis}[
xmin=0,
xmax=1,
ymin=0,
ymax=1.025,
xlabel=Recall,
ylabel=Precision,
legend pos=south west,
legend cell align=right,
]
\addplot+ [mark=none, solid, color=636EFA, line width=1.5pt] table[y=D2DRescore_@_0.5m] {\dataZ};
\addlegendentry{D2DRescore @ \qty{0.5}{\meter}}
\addplot+ [mark=none, solid, color=00CC96, line width=1.5pt] table[y=CircleNMS_@_0.5m] {\dataZ};
\addlegendentry{CircleNMS @ \qty{0.5}{\meter}}
\addplot+ [mark=none, solid, color=EF553B, line width=1.5pt] table[y=GossipNet3D_@_0.5m] {\dataZ};
\addlegendentry{GossipNet3D @ \qty{0.5}{\meter}}
\addplot+ [mark=none, dashed, color=636EFA, line width=1.5pt] table[y=D2DRescore_@_4.0m] {\dataZ};
\addlegendentry{D2DRescore @ \qty{4.0}{\meter}}
\addplot+ [mark=none, dashed, color=00CC96, line width=1.5pt] table[y=CircleNMS_@_4.0m] {\dataZ};
\addlegendentry{CircleNMS @ \qty{4.0}{\meter}}
\addplot+ [mark=none, dashed, color=EF553B, line width=1.5pt] table[y=GossipNet3D_@_4.0m] {\dataZ};
\addlegendentry{GossipNet3D @ \qty{4.0}{\meter}}
\end{axis}
\end{tikzpicture}

%% file: sections/conclusion_and_outlook.tex
\section{Conclusion and Outlook}\label{sec:conclusion_and_outlook}
We present GossipNet3D and D2D-Rescore as learned filtering methods for refining dense 3D object detections using only detection-level features.
Both improve over heuristic \gls{nms} in \gls{map}, \gls{nds}, and true positive quality, with the largest gains for rare or small object classes.
D2D-Rescore achieves competitive accuracy while being more efficient in memory and latency, whereas GossipNet3D remains the strongest in \gls{map}.
Ablations showed that metric-aware matching outperforms Hungarian matching and that refined scores are well-ranked, enabling top-$K$ reduction with minimal loss. 
As these modules rely on a universal detection interface, they are inherently detector-agnostic.
Future work will verify this generalization across further backbones, explore extensions to full box refinement, and analyze why neighborhood-constrained aggregation outperforms unconstrained attention.

%% file: references.bib
@inproceedings{yin2021center,
  title={Center-based 3d object detection and tracking},
  author={Yin, Tianwei and Zhou, Xingyi and Kr\"ahenb\"uhl, Philipp},
  booktitle={Proceedings of the IEEE/CVF conference on computer vision and pattern recognition},
  year={2021},
  language={english}
}

@inproceedings{lang2019pointpillars,
  title={Pointpillars: Fast encoders for object detection from point clouds},
  author={Lang, Alex H and Vora, Sourabh and Caesar, Holger and Zhou, Lubing and Yang, Jiong and Beijbom, Oscar},
  booktitle={Proceedings of the IEEE/CVF conference on computer vision and pattern recognition},
  year={2019},
  language={english}
}

@inproceedings{bodla2017soft,
  title={Soft-NMS--improving object detection with one line of code},
  author={Bodla, Navaneeth and Singh, Bharat and Chellappa, Rama and Davis, Larry S},
  booktitle={Proceedings of the IEEE international conference on computer vision},
  year={2017},
  language={english}
}

@inproceedings{hosang2017learning,
  title={Learning non-maximum suppression},
  author={Hosang, Jan and Benenson, Rodrigo and Schiele, Bernt},
  booktitle={Proceedings of the IEEE conference on computer vision and pattern recognition},
  year={2017},
  language={english}
}

@inproceedings{hu2018relation,
  title={Relation networks for object detection},
  author={Hu, Han and Gu, Jiayuan and Zhang, Zheng and Dai, Jifeng and Wei, Yichen},
  booktitle={Proceedings of the IEEE conference on computer vision and pattern recognition},
  year={2018},
  language={english}
}

@inproceedings{ding2022end,
  title={End-to-end single shot detector using graph-based learnable duplicate removal},
  author={Ding, Shuxiao and Rehder, Eike and Schneider, Lukas and Cordts, Marius and Gall, Juergen},
  booktitle={DAGM German Conference on Pattern Recognition},
  year={2022},
  organization={Springer},
  language={english}
}

@inproceedings{erabati2023li3detr,
  title={Li3detr: A lidar based 3d detection transformer},
  author={Erabati, Gopi Krishna and Araujo, Helder},
  booktitle={Proceedings of the IEEE/CVF Winter conference on applications of computer vision},
  year={2023},
  language={english}
}

@inproceedings{caesar2020nuscenes,
  title={nuscenes: A multimodal dataset for autonomous driving},
  author={Caesar, Holger and Bankiti, Varun and Lang, Alex H and Vora, Sourabh and Liong, Venice Erin and Xu, Qiang and Krishnan, Anush and Pan, Yu and Baldan, Giancarlo and Beijbom, Oscar},
  booktitle={Proceedings of the IEEE/CVF conference on computer vision and pattern recognition},
  year={2020},
  language={english}
}

@inproceedings{carion2020end,
  title={End-to-end object detection with transformers},
  author={Carion, Nicolas and Massa, Francisco and Synnaeve, Gabriel and Usunier, Nicolas and Kirillov, Alexander and Zagoruyko, Sergey},
  booktitle={European conference on computer vision},
  year={2020},
  organization={Springer},
  language={english}
}

@article{gahlert2020visibility,
  title={Visibility guided nms: Efficient boosting of amodal object detection in crowded traffic scenes},
  author={G{\"a}hlert, Nils and Hanselmann, Niklas and Franke, Uwe and Denzler, Joachim},
  journal={arXiv preprint arXiv:2006.08547},
  year={2020},
  language={english}
}

@article{tancik2020fourier,
  title={Fourier features let networks learn high frequency functions in low dimensional domains},
  author={Tancik, Matthew and Srinivasan, Pratul and Mildenhall, Ben and Fridovich-Keil, Sara and Raghavan, Nithin and Singhal, Utkarsh and Ramamoorthi, Ravi and Barron, Jonathan and Ng, Ren},
  journal={Advances in neural information processing systems},
  volume={33},
  year={2020},
  language={english}
}

@inproceedings{wang2023dsvt,
  title={Dsvt: Dynamic sparse voxel transformer with rotated sets},
  author={Wang, Haiyang and Shi, Chen and Shi, Shaoshuai and Lei, Meng and Wang, Sen and He, Di and Schiele, Bernt and Wang, Liwei},
  booktitle={Proceedings of the IEEE/CVF Conference on Computer Vision and Pattern Recognition},
  year={2023},
  language={english}
}

@inproceedings{wang2022detr3d,
  title={Detr3d: 3d object detection from multi-view images via 3d-to-2d queries},
  author={Wang, Yue and Guizilini, Vitor Campagnolo and Zhang, Tianyuan and Wang, Yilun and Zhao, Hang and Solomon, Justin},
  booktitle={Conference on robot learning},
  year={2022},
  organization={PMLR},
  language={english}
}

@article{symeonidis2023neural,
  title={Neural attention-driven non-maximum suppression for person detection},
  author={Symeonidis, Charalampos and Mademlis, Ioannis and Pitas, Ioannis and Nikolaidis, Nikos},
  journal={IEEE transactions on image processing},
  volume={32},
  year={2023},
  publisher={IEEE},
  language={english}
}

@inproceedings{liu2023petrv2,
  title={Petrv2: A unified framework for 3d perception from multi-camera images},
  author={Liu, Yingfei and Yan, Junjie and Jia, Fan and Li, Shuailin and Gao, Aqi and Wang, Tiancai and Zhang, Xiangyu},
  booktitle={Proceedings of the IEEE/CVF international conference on computer vision},
  year={2023},
  language={english}
}

@inproceedings{yang2023bevformerv2,
  title={Bevformer v2: Adapting modern image backbones to bird's-eye-view recognition via perspective supervision},
  author={Yang, Chenyu and Chen, Yuntao and Tian, Hao and Tao, Chenxin and Zhu, Xizhou and Zhang, Zhaoxiang and Huang, Gao and Li, Hongyang and Qiao, Yu and Lu, Lewei and others},
  booktitle={Proceedings of the IEEE/CVF conference on computer vision and pattern recognition},
  year={2023},
  language={english}
}

@inproceedings{wang2023internimage,
  title={Internimage: Exploring large-scale vision foundation models with deformable convolutions},
  author={Wang, Wenhai and Dai, Jifeng and Chen, Zhe and Huang, Zhenhang and Li, Zhiqi and Zhu, Xizhou and Hu, Xiaowei and Lu, Tong and Lu, Lewei and Li, Hongsheng and others},
  booktitle={Proceedings of the IEEE/CVF conference on computer vision and pattern recognition},
  year={2023},
  language={english}
}

@article{lyu2022rtmdet,
  title={Rtmdet: An empirical study of designing real-time object detectors},
  author={Lyu, Chengqi and Zhang, Wenwei and Huang, Haian and Zhou, Yue and Wang, Yudong and Liu, Yanyi and Zhang, Shilong and Chen, Kai},
  journal={arXiv preprint arXiv:2212.07784},
  year={2022},
  language={english}
}
